\documentclass[10pt,twocolumn,letterpaper]{article}

\usepackage[pagenumbers]{cvpr} 

\usepackage{amsmath}
\usepackage{amssymb}
\usepackage{algorithm}
\usepackage{algorithmicx}
\usepackage{algpseudocode}     

\usepackage{booktabs}
\usepackage{multirow}
\usepackage{makecell}
\usepackage{array}
\usepackage{pifont} 

\usepackage{subcaption}  
\usepackage{caption}
\definecolor{cvprblue}{rgb}{0.21,0.49,0.74}
\usepackage[pagebackref,breaklinks,colorlinks,allcolors=cvprblue]{hyperref}

\def\paperID{} 
\def\confName{CVPR}
\def\confYear{2026}

\title{GIFSplat: Generative Prior-Guided Iterative Feed-Forward 3D Gaussian Splatting from Sparse Views}

\author{Tianyu Chen$^{1}$ \quad Wei Xiang$^{1}$\thanks{Corresponding author.} \quad Kang Han$^{1}$ \quad Yu Lu$^{1}$ \quad Di Wu$^{1}$ \quad Gaowen Liu$^{2}$ \quad Ramana Rao Kompella$^{2}$ \\
{\normalsize $^1$La Trobe University, Melbourne, VIC 3086, Australia}\\
{\normalsize $^2$Cisco Research, San Jose, CA, USA}\\
{\tt\small \{t.chen, w.xiang, k.han, y.lu, d.wu\}@latrobe.edu.au,}
{\tt\small \{gaoliu, rkompell\}@cisco.com}
}

\begin{document}

\maketitle
\begin{abstract}
Feed-forward 3D reconstruction offers substantial runtime advantages over per-scene optimization, which remains slow at inference and often fragile under sparse views. However, existing feed-forward methods still have potential for further performance gains, especially for out-of-domain data, and struggle to retain second-level inference time once a generative prior is introduced. These limitations stem from the one-shot prediction paradigm in existing feed-forward pipeline: models are strictly bounded by capacity, lack inference-time refinement, and are ill-suited for continuously injecting generative priors. We introduce GIFSplat, a purely feed-forward iterative refinement framework for 3D Gaussian Splatting from sparse unposed views. A small number of forward-only residual updates progressively refine current 3D scene using rendering evidence, achieve favorable balance between efficiency and quality.  Furthermore, we distill a frozen diffusion prior into Gaussian-level cues from enhanced novel renderings without gradient backpropagation or ever-increasing view-set expansion, thereby enabling per-scene adaptation with generative prior while preserving feed-forward efficiency. Across DL3DV, RealEstate10K, and DTU, GIFSplat consistently outperforms state-of-the-art feed-forward baselines, improving PSNR by up to +2.1 dB, and it maintains second-scale inference time without requiring camera poses or any test-time gradient optimization.
\end{abstract}

\section{Introduction}
\label{sec:intro}


Reconstructing 3D scenes from multi-view images is a fundamental task in computer vision, perception, and computer graphics, underpinning applications in robotic perception, AR/VR, and high-fidelity 3D content creation. 
3D reconstruction has rapidly progressed along two paradigms: per-scene optimization and feed-forward. Per-scene optimization methods \cite{mildenhall2021nerf,kerbl20233d,yu2024mip} can achieve high-fidelity reconstructions by iteratively minimizing photometric objectives on a single scene, but their inference requires long test-time gradient optimization, which limits scalability in real applications. In addition, while such pipelines benefit from iterative gradient feedback, they often struggle under sparse view inputs and do not effectively incorporate external data priors beyond inductive bias and observed samples, leaving the rich knowledge contained in large-scale 3D data underutilized.

\begin{figure}[t]
  \centering
  \includegraphics[width=0.95\columnwidth]{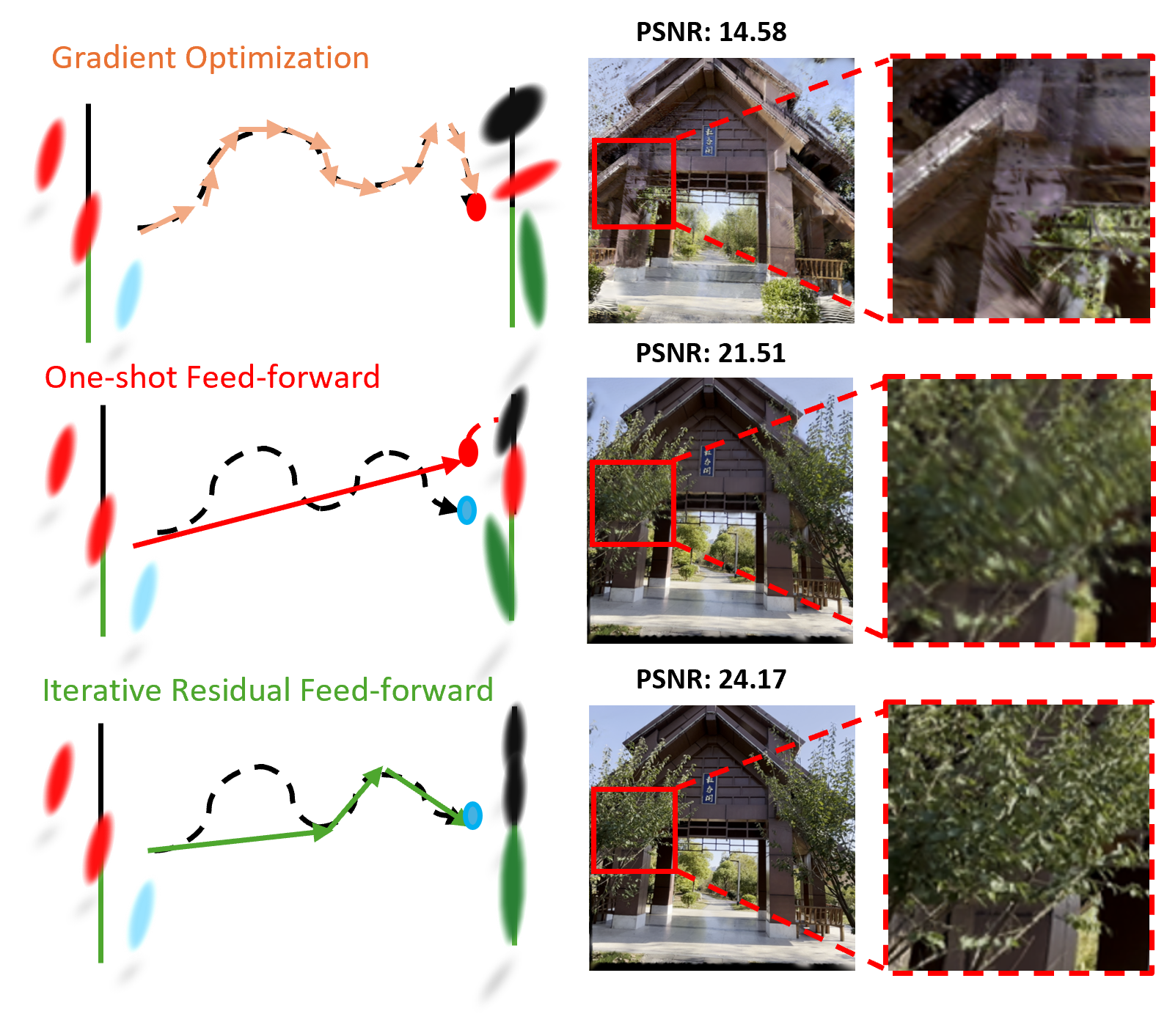}
  \caption{\textbf{Conceptual comparison of reconstruction paradigms.} Gradient optimization performs thousands updates, incurring heavy test-time cost, often achieving high quality in dense-view scenarios but struggling in sparse-view scenarios; One-shot feed-forward \cite{jiang2025anysplat} is efficient but leaves noticeable artifacts; Our iterative residual feed-forward scheme keeps feed-forward efficiency and achieves higher reconstruction quality without test-time gradient backpropagation.}
  \label{fig:1}
\end{figure}

In contrast, recent feed-forward approaches ~\cite{wang2024dust3r,leroy2024grounding,yang2025fast3r,wang2025vggt} estimate 3D attributes from multi-view images in a single forward pass, incorporating multi-view interactions through an attention mechanism within Vision Transformers and employing dedicated attribute prediction heads, achieving millisecond- to second-scale inference time.
However, such one-shot prediction introduces two key limitations:
(1) its performance is constrained by model capacity, leading to lower fidelity in complex scenes; and
(2) it lacks scene-specific refinement, leaving residual errors uncorrected even in large-scale models \cite{wang2025vggt}.
Consequently, the core challenge remains: \emph{how to achieve feed-forward efficiency while enabling adaptation and continuous refinement to each scene.}



However, generating artifact-free novel views remains challenging in complex scenes, particularly when view coverage is sparse \cite{kerbl20233d}. Recent studies \cite{wu2025difix3d+,wu2025genfusion} tend to address this limitation by integrating generative priors but require long-time per-scene optimization.
They alternatively optimize 3D Gaussians and refine rendered views with diffusion models, and these refined views are then fed back into the training set for further optimization, forming a feedback loop that gradually improves reconstruction quality. 
Yet this iterative loop is \emph{ill-suited} for existing feed-forward pipelines. As the continuously growing set of synthesized views increases both time and memory complexity, which are further exacerbated by the expensive self-attention operations in ViT-based backbones \cite{sharir2021image} adopted by recent feed-forward methods \cite{charatan2024pixelsplat, chen2024mvsplat, xu2025depthsplat, ye2024no, jiang2025anysplat}. Moreover, these methods reconstruct 3D scene directly from multi-view images rather than updating an existing scene representation, making iterative enhancement feedback infeasible, let alone step-wise fusion of generative priors. This raises an essential question: \emph{how to refine a 3D state in a purely feed-forward manner, leveraging both observations and generative priors while keeping second-scale inference runtime?}

To tackle this challenge, we propose GIFSplat, an iterative feed-forward 3D Gaussian Splatting framework that leverages the strengths of both feed-forward and optimization-based methods. The first stage performs a fast feed-forward pass to produce a reliable initialization, and the second stage refines an initial prediction through multi-step forward-only residual updates guided by scene evidence and diffusion-enhanced cues. 
GIFSplat maintains the speed of feed-forward pipeline while enabling scene-adaptive refinement and reliable prior injection, bridging the gap between one-shot feed-forward methods and optimization-based pipelines.
At each step, GIFSplat uses the current Gaussian state and discrepancy cues to predict residual updates.
This allows continuous improvement without test-time gradient backpropagation, with memory and runtime scaling linearly in the number of refinement steps. A conceptual comparison of paradigms is shown in ~\cref{fig:1}.
Rather than continuously extending supervised image sets with enhancing rendered images then re-optimizing the scene, we apply a diffusion enhancer \cite{wu2025difix3d+} to novel rendered images, extract enhancement deltas, and convert them into prior guidance cue. This approach brings reliable priors without view explosion and followed heavy computation.

The primary contributions of this work are as follows:
\begin{itemize}
\item We present an iterative feed-forward update mechanism for 3D Gaussian Splatting that refines a fixed set of Gaussians via multi-step, forward-only residual updates, enabling scene-specific refinement without any test-time gradient descent.

\item We introduce a generative prior fusion mechanism that distills a frozen diffusion prior into lightweight Gaussian-level discrepancy cues from enhanced novel renderings, injecting generative information into the iterative updates without backpropagation or continuously expanding the set of reference views.

\item We demonstrate consistent performance gains across DL3DV, RealEstate10K, and DTU datasets, under varying view-overlap settings, showing that our iterative feed-forward model (IFSplat) and generative-prior variant (GIFSplat) both outperform recent feed-forward reconstruction methods in sparse-view and cross-domain scenarios.
\end{itemize}



\section{Related Work}
\label{sec:formatting}

\subsection{3D Reconstruction from Multi-view Images}
Classical per-scene optimization pipelines, such as NeRF~\cite{mildenhall2021nerf}, 3D Gaussian Splatting (3DGS)~\cite{kerbl20233d} and its variants~\cite{kerbl20233d, yu2024mip, muller2022instant}, achieve high visual fidelity but require thousands of test-time gradient steps and meet substantial performance drop in sparse-view setting, limiting practicality in real applications. To bypass slow SfM/MVS pipelines and strong pose dependence while improving quality in sparse view settings, recent feed-forward approaches infer 3D structure directly from images.
\begin{figure*}[htbp]
  \centering
  \includegraphics[width=0.98\linewidth]{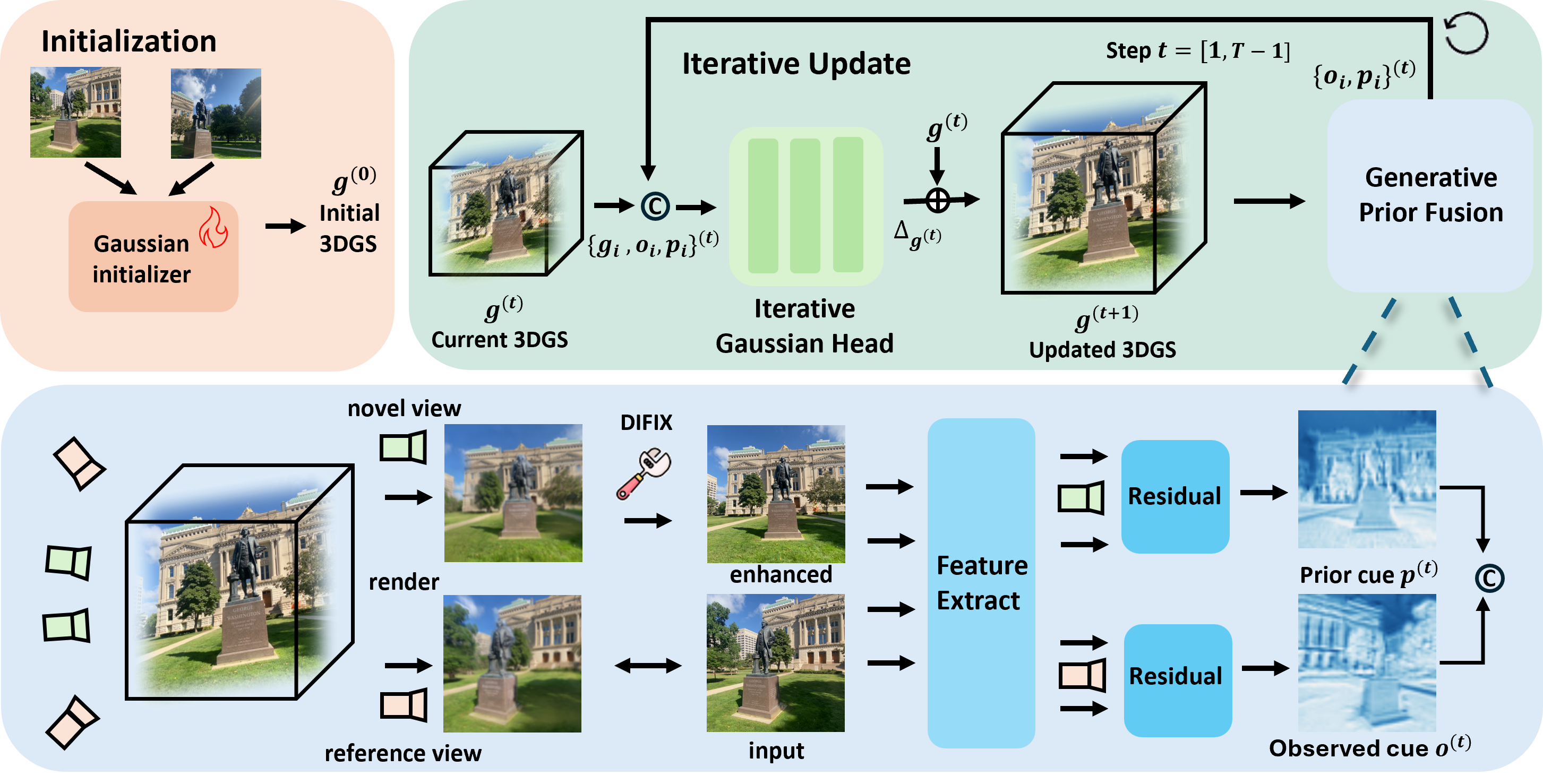}
  \caption{\textbf{Overview of GIFSplat.} Our framework consists of a Gaussian initializer, an iterative Gaussian head, and a generative prior fusion module. The initializer takes sparse input views and predicts camera parameters and initial 3DGS $g_0$. The iterative head then refines the Gaussians over several forward-only steps by updating per-Gaussian parameters $g_i$ using residual corrections $\Delta g$ predicted from the concatenated state and cues $\{g_i, \mathbf{o}_i, \mathbf{p}_i\}$. At each step, we render reference and novel views, compute observation evidence $\mathbf{o}_i$ from feature differences between input and rendered views, and derive generative prior cues $\mathbf{p}_i$ by enhancing the renderings with a frozen diffusion model DIFIX and taking feature-space residuals. These Gaussian-level signals are fed back into the iterative head to progressively improve the 3DGS, particularly in under-constrained regions.}
  \label{fig:overview}
\end{figure*}

DUSt3R~\cite{wang2024dust3r} predicts dense pointmaps from uncalibrated pairs, removing explicit matching and bundle adjustment, though it still requires pairwise fusion for larger image sets~\cite{leroy2024grounding}. VGGT~\cite{wang2025vggt} treats multi-view geometry as a single large-scale model task, jointly predicting cameras, depth, and pointmaps in one pass. For the 3DGS prediction, NoPoSplat and FLARE~\cite{ye2024no, zhang2025flare} explore pose-free sparse-view lifting to produce 3D Gaussians, and AnySplat~\cite{jiang2025anysplat} jointly estimates cameras and Gaussians with redundancy consolidation for unconstrained inputs. Concurrently, concurrent work iLRM~\cite{kang2025ilrm} explicitly unrolls the iterative refinement into multiple transformer layers with non-shared parameters, where each layer performs cross-attention with the reference-view images to inject features and apply one update. As a result, increasing the number of refinement steps makes the parameter count grow linearly with the depth, and the number of refinement steps is fixed at inference time for a given model. In contrast, we adopt a single weight-shared refinement module that keeps the parameter count constant while allowing the number of iterations to be flexibly adjusted during inference. While concurrent work ReSplat \cite{xu2025resplat} performs iterative feed-forward refinement based on observational errors, it still relies on known camera poses. In contrast, our pose-free framework uniquely addresses under-constrained sparse views by injecting on-the-fly generative diffusion priors directly into the update loop.

Overall, this line shifts reconstruction from slow optimization to fast forward inference. However, most remain one-shot predictors and lack reliable scene-adaptive refinement mechanisms, precisely the gap addressed by our iterative feed-forward refinement framework.


\subsection{Generative Priors for 3D Reconstruction}
The aforementioned methods perform well when novel viewpoints are close to the input views, but their performance drops markedly under sparse or low-coverage inputs, often producing artifacts in under-constrained regions. Generative models provide strong data-driven priors for under-constrained regions. Early approaches such as DreamFusion~\cite{poole2022dreamfusion} distill 2D diffusion into 3D but rely on lengthy per-scene optimization. Later work seeks tighter integration: multi-view diffusion model~\cite{szymanowicz2025bolt3d}, and rectified-flow formulations~\cite{wang2024splatflow} enable efficient 3DGS generation rather than 3D reconstruction, while limited to simple scene.

Diffusion-enhanced 3D reconstruction pipelines ~\cite{yu2024viewcrafter, wu2025genfusion,liu20243dgs,fischer2025flowr} render provisional views, enhance them via generative priors, and feed them back for improved coarse 3D model, both of them need multi-step generation. Difix3D+~\cite{wu2025difix3d+} compresses diffusion into a single-step enhancer for faster enhancement of rendered view. While above techniques significantly boost perceptual quality, they still depend on optimization-based regimes rather than updating in a feed-forward manner, meaning they require time-consuming optimization runs.

These developments highlight the importance of generative priors for sparse-view and out-of-distribution scenes, but also reveal the need for a framework that \emph{retains feed-forward efficiency while enabling stable, scene-adaptive refinement with generative prior}. Our work pursues this direction through iterative forward-only residual refinement with prior injection.

\section{Method}


\subsection{Problem Definition and Framework Overview}
\label{sec:problem_definition}

\paragraph{Problem definition} 
Given a set of uncalibrated multi-view image observations
$\mathcal{V}=\{(I_m)\}_{m=1}^{\mathcal{M}}$,
we represent a scene with 3D Gaussian Splatting (3DGS)
$\mathcal{G}=\{g_i\}_{i=1}^{N}$, where each Gaussian
$g_i=(\mathbf{x}_i,\mathbf{s}_i,\mathbf{r}_i,\mathbf{c}_i,\alpha_i)$
contains position, scale, orientation, color, and opacity. Our goal is to recover $\mathcal{G}$ that is photometrically consistent with $\mathcal{V}$,
while enabling fast inference and robustness under domain shift. A differentiable rasterizer $\mathcal{R}$ produces a rendering
$R_m=\mathcal{R}(\mathcal{G};\Pi_m)$ for camera $\Pi_m$.

\paragraph{Framework overview} 
To combine feed-forward efficiency with scene-specific refinement and efficient use of generative prior, our framework consists of three components (in ~\cref{fig:overview}): 
We remove voxelization module in \cite{jiang2025anysplat} and fine-tune it as initializer $F_\phi$ of our framework, which first predicts camera poses and an initial 3DGS $\mathcal{G}^{(0)}$ from $\mathcal{V}$. On top of this, an iterative residual Gaussian head $U_\theta$ applies $T$ forward-only updates to refine the 3D Gaussians (in ~\cref{sec:iter_gaussian_head}). Meanwhile, a generative prior fusion mechanism converts diffusion-enhanced renderings into Gaussian-level cues that guide these updates (in ~\cref{sec:prior_fusion}).
This decomposition allows us to retain the speed of feed-forward inference while still with scene-adaptive refinement.

\subsection{Iterative Gaussian Head}
\label{sec:iter_gaussian_head}

\begin{figure}[t]
  \centering
  \includegraphics[width=\columnwidth]{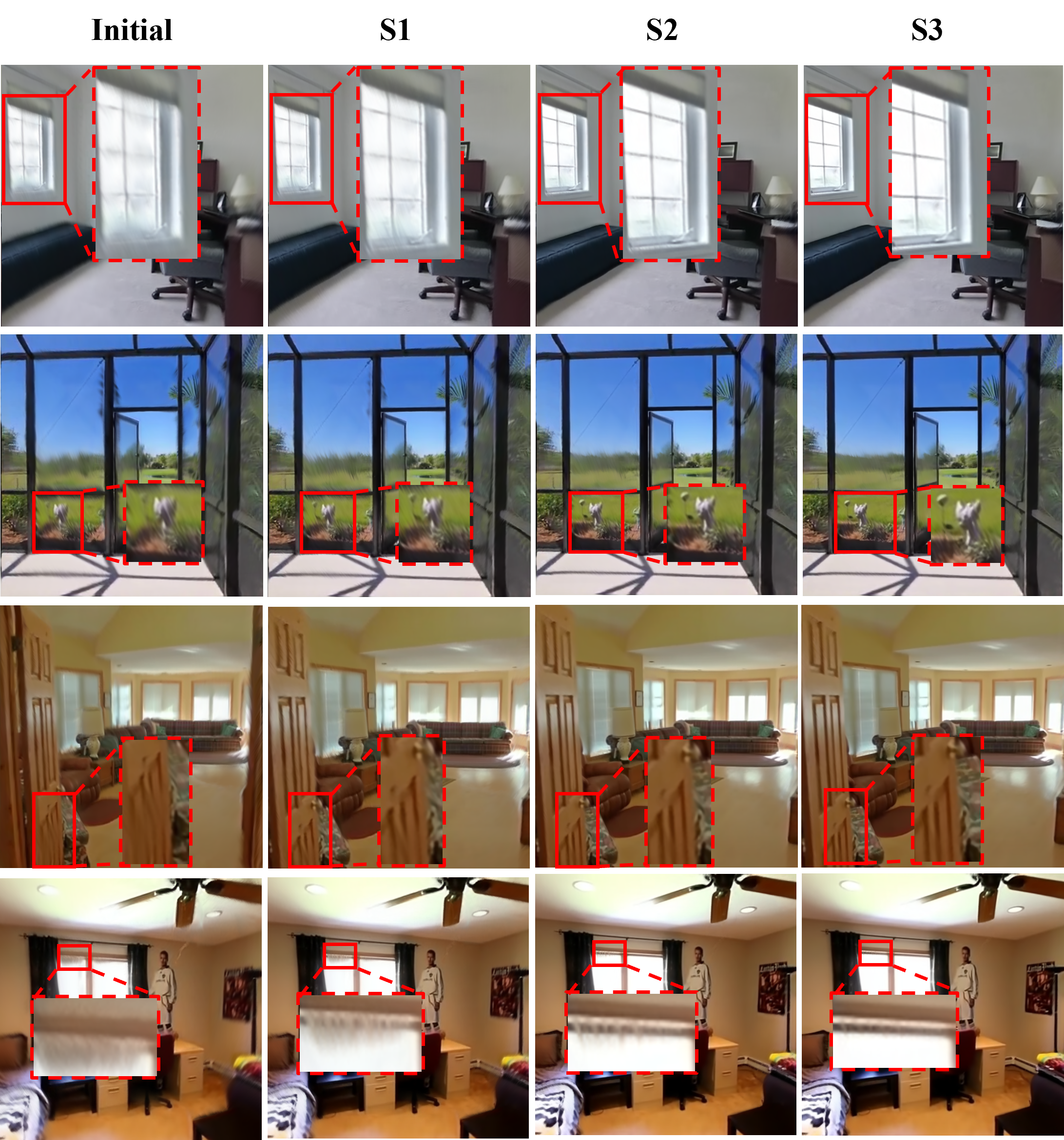}
  \caption{\textbf{Visualizing iterative residual refinement.} Starting from the initial 3DGS prediction (left column), our iterative Gaussian head progressively refines geometry and appearance over three forward-only steps (S1–S3). The zoomed regions highlighted by red dashed boxes show reduced blur, sharper edges, and fewer artifacts as the iteration proceeds, illustrating how the proposed updates gradually improve the scene representation without test-time gradient backpropagation.}
  \label{fig:IR}
\end{figure}

Instead of one-shot prediction, we adopt an iterative feed-forward refinement:
starting from an initialization $\mathcal{G}^{(0)}$, we apply $T$ forward update steps to obtain
$\mathcal{G}^{(T)}$, without test-time gradient backpropagation or long per-scene optimization. In short, existing paradigms satisfy only partial property at a time:
feed-forward efficiency, refinement ability, and prior utilization. We seek a formulation satisfying all three. ~\cref{fig:IR} illustrates the process of iterative refinement.


Given the formulation above, we now detail how the iterative gaussian head enables
feed-forward scene refinement without test-time gradient backpropagation.
Existing optimization-based methods refine 3D Gaussians through long-time gradient
descent, whereas one-shot feed-forward models lack the ability to absorb new
evidence. Our goal is to combine their strengths: efficient inference while still allowing scene-adaptive refinement.
Under \emph{no} test-time gradients, the iterative head transforms the initial
3DGS $\mathcal{G}^{(0)}$ into $\mathcal{G}^{(T)}$ by predicting per-Gaussian residual $\{g_i\}^{(T)}$ across $T$ forward passes, reducing rendering–observation discrepancies and optionally incorporating prior cues. Equivalently, it forward-approximates minimizing
$\|\psi(I_m)-\psi(\mathcal{R}(\mathcal{G};\Pi_m))\|$, where $\psi$ is a
frozen image feature extractor and $\mathcal{R}$ is a differentiable renderer, more details are stated in Appendix.

At step $t$, the rendered views are processed by $\psi(\cdot)$ to compute feature differences $\mathcal{O}_m^{(t)}=\psi(I_m)-\psi(R_m^{(t)})$. Pixel cues are pooled to Gaussians via soft assignment weights $w_i(u)$, producing $\mathbf{o}_i^{(t)}$. 
\begin{equation}
\mathbf{o}_i^{(t)} \leftarrow
\frac{\sum_{m\in\mathcal{S}^{(t)}}\sum_{u} w_i(u)\,O_m^{(t)}(u)}
     {\sum_{m\in\mathcal{S}^{(t)}}\sum_{u} w_i(u)+\varepsilon}.
\end{equation}

The iterative gaussian head predicts residuals
$\Delta \mathcal{G}^{(t)} = \Delta \{g_i^{(t)}\}_{i=1}^{N}=(\Delta\mathbf{x}_i^{(t)},\Delta\mathbf{s}_i^{(t)},
\Delta\mathbf{c}_i^{(t)},\Delta\alpha_i^{(t)})$ and update the current 3DGS as
\begin{equation}
    \Delta \mathcal{G}^{(t+1)} \leftarrow U_\theta\big(\,[\mathcal{G}^{(t)} \,\|\, \{o_i\}^{(t)}]\,\big),
\end{equation}
\begin{equation}
    \mathcal{G}^{(t+1)} \leftarrow \mathcal{G}^{(t)}+\Delta \mathcal{G}^{(t)},\quad t=0,\dots,T-1.
\end{equation}

This design preserves feed-forward efficiency while providing a lightweight refinement signal on top of the one-shot initialization.

Previous feed-forward approaches~\cite{charatan2024pixelsplat,chen2024mvsplat,ye2024no}
adopt pixel-aligned Gaussians, coupling splats to image grids. Such alignments are
computationally heavy, misaligned with true 3D neighborhoods, and prone to
over-densifying smooth areas while under-representing detailed or occluded
regions. To ensure refinement respects 3D geometry, we convert pixel-aligned
Gaussians to point-based Gaussians and project correspondent pretrained features by camera. In addition, we use window attention~\cite{zhao2021point} to model local relationships of 3D gaussians efficiently.


For training, we adopt the architecture from~\cite{jiang2025anysplat} as initializer but remove voxelization. It is fine-tuned partially with ground-truth poses, and we compute losses on novel views. The iterative head is unrolled for $T{=}3$ steps with $|\mathcal{S}^{(t)}|$ matching inference. We supervise with MSE and perceptual loss~\cite{zhang2018unreasonable}.

\subsection{Generative Prior Fusion}
\label{sec:prior_fusion}

\begin{figure}[t]
  \centering
  \includegraphics[width=0.95\columnwidth]{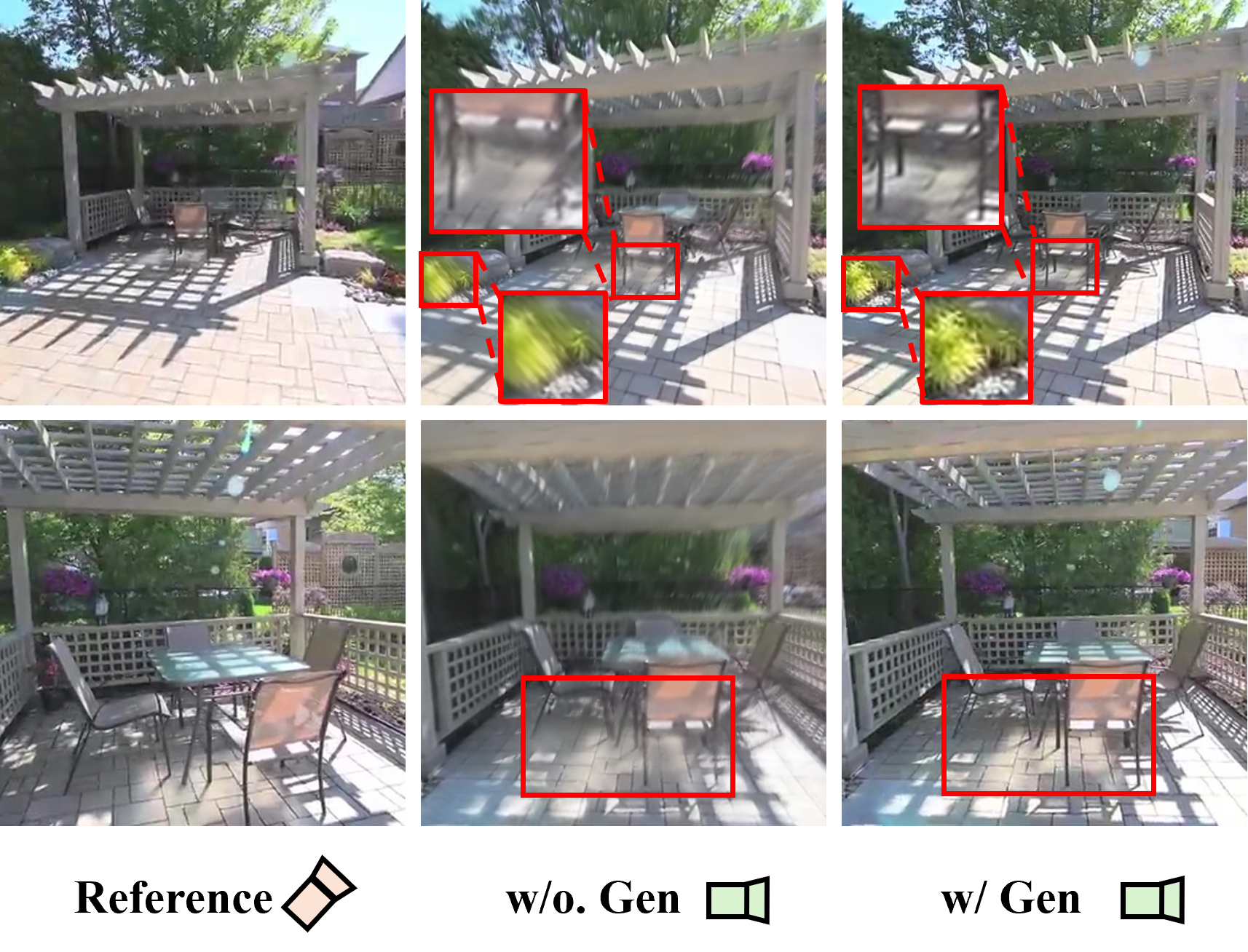}
  \caption{\textbf{Generative prior fusion.} Given sparse input views (left), our feed-forward 3DGS first produces a rendered view (middle). A frozen diffusion-based enhancer then refines this rendering into an enhanced image (right) with sharper textures and richer details, which is converted into Gaussian-level prior cues for subsequent residual updates.}
  \label{fig:difix}
\end{figure}

When observation-only cues are weak (e.g., sparse views or domain shift), generative prior can provide complementary high-frequency appearance. Beyond observation-only refinement, we inject a frozen diffusion-based generative prior. To avoid test-time optimization cost, we adopt a conservative design: the diffusion enhancer is frozen and without using of gradient when inference. Rather than re-optimizing the scene by extending the supervised image set from enhanced views, we enhance intermediate renderings and distill the enhancement into Gaussian-level cues for the next residual update, improving detail fidelity while preserving feed-forward efficiency, as illustrated in \ref{fig:difix}.

For each $m\in\mathcal{S}^{(t)}$, we enhance the current rendering using a diffusion-based enhancer $\mathcal{E}_\phi$ and pretrained model Difix \cite{wu2025difix3d+}: $\tilde{R}_m^{(t)}=\mathcal{E}_\phi\!\big(R_m^{(t)}\big).$
We compute a prior difference in feature space: $P_m^{(t)}=\psi\!\big(\tilde{R}_m^{(t)}\big)-\psi\!\big(R_m^{(t)}\big).$ Then pooling to Gaussians mirrors the evidence pooling to get $\mathbf{p}_i^{(t)}$. These cues regularize forward updates of $U_\theta$ under domain shift and provide high-frequency hints in under-constrained regions. The generative prior cues are similar to observed cues.
\begin{equation}
\mathbf{p}_i^{(t)} \;=\;
\frac{\sum_{m\in\mathcal{S}^{(t)}}\sum_{u} w_i(u)\,P_m^{(t)}(u)}
     {\sum_{m\in\mathcal{S}^{(t)}}\sum_{u} w_i(u)+\varepsilon}.
\end{equation}

We fuse observation-driven evidence and prior cues by concatenation at the gaussian level.
\begin{equation}
\Delta \mathcal{G}_i^{(t)} \;=\; U_\theta\big(\,[g_i^{(t)} \,\|\, \{o_i\}^{(t)} \,\|\, \{p_i\}^{(t)}]\,\big),
\end{equation}
\begin{equation}
    \mathcal{G}^{(t+1)}=\mathcal{G}^{(t)}+\Delta\mathcal{G}^{(t)}.,\quad t=0,\dots,T-1,
\end{equation}
where $g_i^{(t)}$ denotes the Gaussian-state including current per-gaussian parameters and corresponding features, $\{o_i\}^{(t)}$ the observation cues are distilled from rendered–image feature differences, and $\{p_i\}^{(t)}$ the prior cues distilled from diffusion-enhanced renderings. We do not implement gradient backpropagation through single-step diffusion model $\mathcal{E}_\phi$, the prior serves only as a forward cue. 

Our framework enables feed-forward efficiency, scene-adaptive refinement, and use of generative priors without ever-increasing view synthesis or test-time gradient updates, scales roughly linearly with $T$ and the per-iteration view budget, and improves robustness under domain shift via the refinement by observed cue and generative prior.

\subsection{Loss Functions}
\label{sec:loss}

We train in two stages, applying geometric
distillation $\mathcal{L}_{\mathrm{dist}}$ as \cite{jiang2025anysplat} and photometric loss in novel view during Stage~I to bootstrap a stable initialization.

In stage 1 initialization ($t{=}0$), we supervise the first forward prediction with a pixel reconstruction loss $\mathcal{L}_{\mathrm{rec}}$
\begin{equation}
\mathcal{L}_{\mathrm{rec}} =
\lambda_{\mathrm{rgb}}\|I_m - R_m\|_2.
\end{equation}
\begin{equation}
    \mathcal{L}_{\mathrm{stage1}}
=\sum_{m\in\mathcal{M}}\big(
\mathcal{L}_{\mathrm{rec}}+\mathcal{L}_{\mathrm{dist}}
\big).
\end{equation}

In stage 2 refinement ($t{>}0$), we unroll $T$ forward-only refinement steps and supervise each rendering using
the multi-steps render loss $ \mathcal{L}_{\mathrm{stage2}}$. More details are stated in Appendix.
\begin{equation}
    \mathcal{L}_{\mathrm{stage2}}
=\sum_{t=1}^{T}\omega_t\sum_{m\in\mathcal{M}}
\|I_m - R_m^{(t)}\|_2.
\end{equation}

\section{Experiments}
\subsection{Experimental Settings}

\paragraph{Datasets} We train our model on datasets DL3DV \cite{ling2024dl3dv} and RealEstate10K \cite{zhou2018stereo}, which contains in-door and out-door scene. We test in the remained set from training of DL3DV and RealEstate10K, and test the out-of-domain performance on DTU \cite{jensen2014large}.

\newcommand{\w}[1]{\text{w/\ #1}}
\newcommand{\wo}[1]{\text{w/o\ #1}}

\newcommand{\cmark}{\ding{51}} 
\newcommand{\xmark}{\ding{55}} 
\newcolumntype{Y}{>{\centering\arraybackslash}m{2.2cm}}
\newcolumntype{C}[1]{>{\centering\arraybackslash}m{#1}}
\newcolumntype{Y}{>{\raggedright\arraybackslash}p{1.8cm}}

\begin{table*}[htbp]
\centering
\caption{\textbf{Novel view synthesis performance comparison on the RealEstate10K with 2 views as input}. 
Our method largely outperforms pose-free and pose-required methods across all overlap settings, especially for small overlap setting.
The \textbf{best} and \underline{second-best} results are highlighted.}
\label{tab:re10k_nvs}
\renewcommand{\arraystretch}{1.25}
\scriptsize
\resizebox{\linewidth}{!}{%
\begin{tabular}{l l ccc ccc ccc ccc}
\toprule
& \multirow{2}{*}{\textbf{Method}} &
\multicolumn{3}{c}{\textbf{Small}} &
\multicolumn{3}{c}{\textbf{Medium}} &
\multicolumn{3}{c}{\textbf{Large}} &
\multicolumn{3}{c}{\textbf{Average}} \\
\cmidrule(lr){3-5}\cmidrule(lr){6-8}\cmidrule(lr){9-11}\cmidrule(lr){12-14}
& & PSNR$\uparrow$ & SSIM$\uparrow$ & LPIPS$\downarrow$ &
PSNR$\uparrow$ & SSIM$\uparrow$ & LPIPS$\downarrow$ &
PSNR$\uparrow$ & SSIM$\uparrow$ & LPIPS$\downarrow$ &
PSNR$\uparrow$ & SSIM$\uparrow$ & LPIPS$\downarrow$ \\
\midrule
\multirow{4}{*}{\rotatebox{90}{Pose-Required}} 
& pixelNeRF \cite{yu2021pixelnerf} & 18.417 & 0.601 & 0.526 & 19.930 & 0.632 & 0.480 & 20.869 & 0.639 & 0.458 & 19.824 & 0.626 & 0.485 \\
& AttnRend \cite{du2023learning} & 19.151 & 0.663 & 0.368 & 22.532 & 0.763 & 0.269 & 25.897 & 0.845 & 0.186 & 22.664 & 0.762 & 0.269 \\
& pixelSplat \cite{charatan2024pixelsplat} & 20.263 & 0.717 & 0.266 & 23.711 & 0.809 & 0.181 & 27.151 & 0.879 & 0.122 & 23.848 & 0.806 & 0.185 \\
& MVSplat \cite{chen2024mvsplat}   & 20.353 & 0.724 & 0.250 & 23.778 & 0.812 & 0.173 & 27.408 & 0.884 & 0.116 & 23.977 & 0.811 & 0.176 \\
\midrule
\multirow{9}{*}{\rotatebox{90}{Pose-Free}} 
& DUST3R \cite{wang2024dust3r}  & 14.101 & 0.432 & 0.468 & 15.419 & 0.451 & 0.432 & 16.427 & 0.453 & 0.402 & 15.382 & 0.447 & 0.432 \\
& MAST3R \cite{leroy2024grounding}   & 13.534 & 0.407 & 0.494 & 14.945 & 0.436 & 0.451 & 16.028 & 0.444 & 0.418 & 14.907 & 0.431 & 0.452 \\
& Splatt3r \cite{smart2024splatt3r}  & 14.352 & 0.475 & 0.472 & 15.529 & 0.502 & 0.425 & 15.817 & 0.483 & 0.421 & 15.318 & 0.490 & 0.436 \\
& CoPoNeRF \cite{hong2023unifying} & 17.393 & 0.585 & 0.462 & 18.813 & 0.616 & 0.392 & 20.464 & 0.652 & 0.318 & 18.938 & 0.619 & 0.388 \\
& NopoSplat \cite{ye2024no} & 22.514 & 0.784 & 0.210 & 24.899 & 0.839 & 0.160 & 27.411 & 0.883 & 0.119 & 25.033 & 0.838 & 0.160 \\
& FLARE \cite{zhang2025flare}  &  22.468 & 0.763 & 0.198 & 24.872 & 0.836 & 0.162 & 27.012 & 0.881 & 0.123 & 24.891 & 0.831 & 0.161  \\ 
& AnySplat \cite{jiang2025anysplat} &  22.703	& 0.781	& 0.221 &	25.107	& 0.844 &	0.156	& 27.414	& 0.883	& 0.118 & 25.176 & 0.839 & 0.161 \\ 
\cmidrule(lr){2-14}
& \textbf{IFSplat(Ours)} & \underline{24.207} & \underline{0.813} & \underline{0.183} & \underline{26.421} & \underline{0.853} & \underline{0.145} & \underline{27.895} & \underline{0.891} & \underline{0.114} & \underline{26.291} & \underline{0.854} & \underline{0.145} \\
& \textbf{GIFSplat(Ours)} & \textbf{24.617} & \textbf{0.827} & \textbf{0.181} & \textbf{26.714} & \textbf{0.866} & \textbf{0.131} & \textbf{27.987} & \textbf{0.902} & \textbf{0.112} & \textbf{26.559}  & \textbf{0.867} & \textbf{0.138} \\
\bottomrule
\end{tabular}}
\end{table*}


\begin{figure*}[ht]
  \centering
  \includegraphics[width=0.97\linewidth]{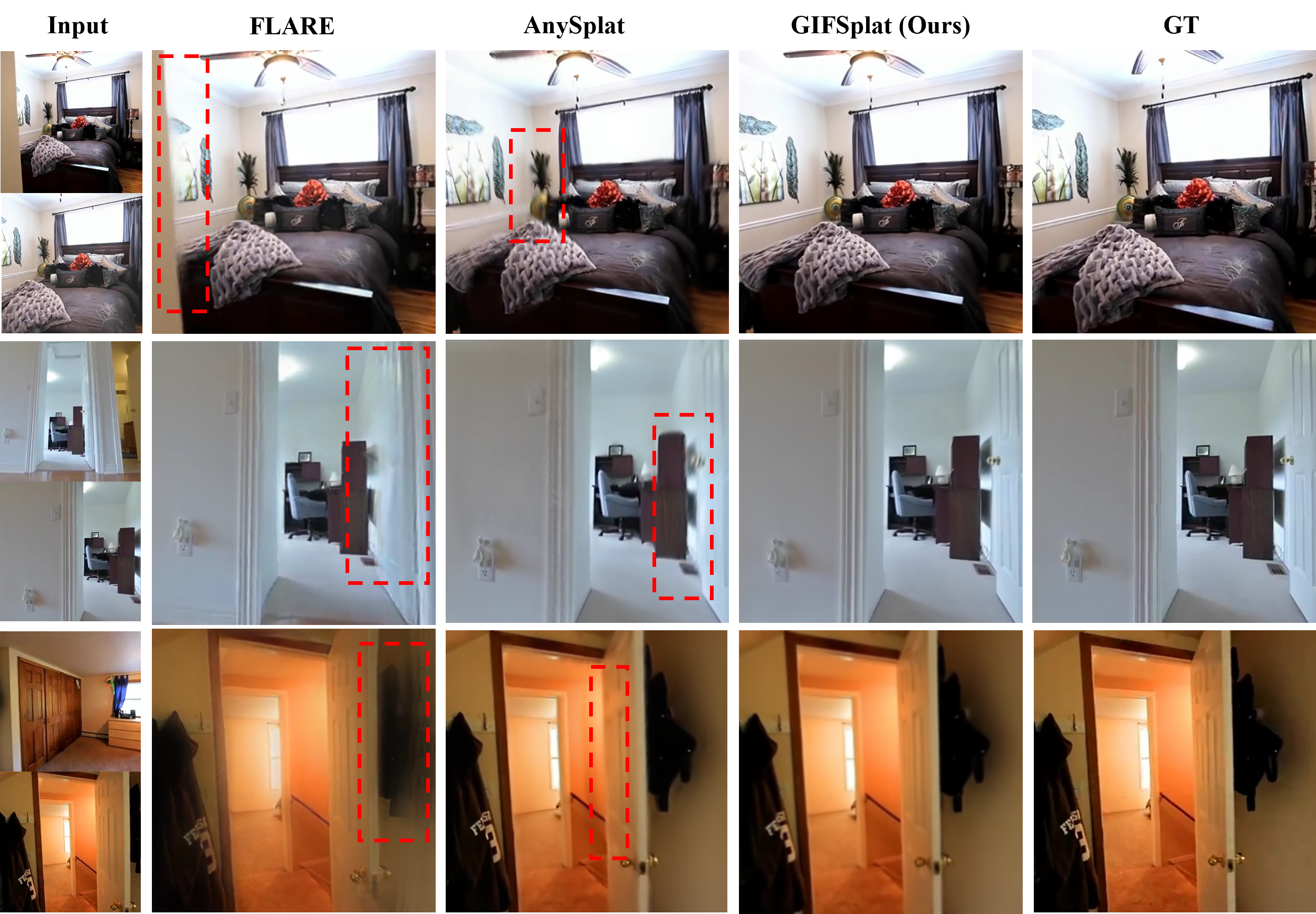}
  \caption{\textbf{Qualitative comparisons on representative indoor scenes from RealEstate10K.}
 Columns: sparse input views, FLARE, AnySplat, our GIFSplat with generative prior, and ground truth (GT). Red dashed boxes highlight that GIFSplat recovers sharper boundaries (e.g., door frames, wall corners), more faithful textures, and fewer artifacts such as texture sticking.}
  \label{fig:re10k}
\end{figure*}


\paragraph{Baselines and metrics} 
To compare with previous methods, we mainly consider three evaluation setups. First, novel view synthesis from 8 input views at 448×448 resolution on DL3DV, where we adopt PixelSplat, MVSplat, FLARE, AnySplat as baselines. Second, we also evaluate on the commonly used 2-view at 256×256 resolution setup on RealEstate10K \cite{zhou2018stereo} with different overlap settings, except for aforementioned baseline, we compare with benchmark in ~\cite{ye2024no}. In addition, we test the generalization ability on DTU using model trained on RealEstate10K. We use PSNR, SSIM, LPIPS as our metric to evaluate the performance of novel view synthesis. More information is provided in the  Appendix.

\paragraph{Experimental settings} We use gsplat \cite{yu2024mip} as differential rasterizer and AdamW as optimizer . We train and evaluate our model on 4 NVIDIA DGX H200 141GB GPU for all experiments. For experiments on DL3DV, we fine-tuned the initializer for 30K steps and trained the iterative gaussians head for 300K steps. For experiments on RealEstate10K, we fine-tuned the initializer for 50K steps and trained the iterative gaussians head for 100K steps. More information is provided in the Appendix. For DL3DV, we follow the FLARE test split and evaluate in 112 unseen scenes (8 views for input, 9 views for evaluation). For RealEstate10K (2 views for input, 3 views for evaluation) and generalization on DTU (2 views for input, 4 views for evaluation) experiments, we all follow the NopoSplat evaluation setting.

\subsection{Comparison with Feed-forward baseline}

\paragraph{2-view evaluation on RealEstate10K}

~\cref{tab:re10k_nvs} and ~\cref{fig:re10k} show the quantitative and qualitative results, respectively. We follow the settings of NopoSplat \cite{ye2024no}, splitting the scene to three parts according to the overlap ratio. Across all setting, our iterative residual model IFSplat get the better results than existing baselines. Moreover, GIFSplat gets the best results with the help of generative prior, preserving fine textures, with fewer blurs than all other models. Notably, some implausible deformation such as the door and wardrobe are rectified by generative prior. 

\paragraph{8-view evaluation on DL3DV}

\begin{figure}[htbp]
  \centering
  \includegraphics[width=\linewidth]{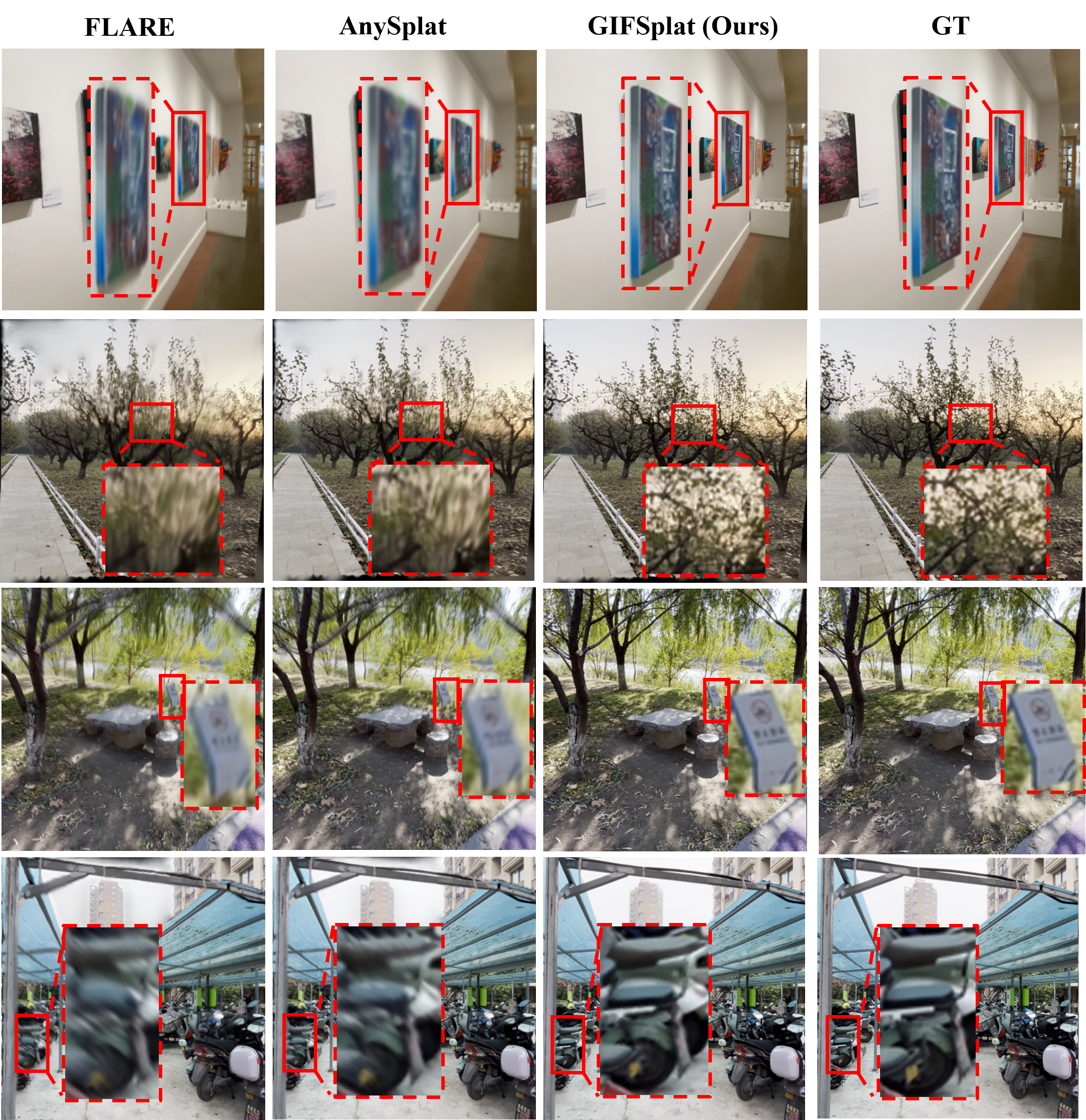}
  \caption{\textbf{Qualitative comparisons on DL3DV}. 
Columns: FLARE, AnySplat, GIFSplat (ours), and GT. 
Across representative scenes, GIFSplat preserves sharper edges and textures while suppressing blur and texture sticking compared with feed-forward baselines. }
\label{fig:DL3DV}
\end{figure}

We evaluate on dataset DL3DV with 8 input views, results are summarized in ~\cref{tab:dl3dv}. To reduce storage, when training the model for 8 views, we downsample 2x the depth and project to 3D space as location of 3D gaussians, and predict other parameters of 3d gaussians.
Across both baselines, GIFSplat consistently outperforms strong feed-forward baselines on all metrics in ~\cref{tab:dl3dv}, aligning with the sharper, artifact-free visuals in ~\cref{fig:DL3DV}.
Notably, our method attains these improvements without relying on camera pose, demonstrates the robustness to sparse views.


\begin{table}[htbp]
\centering
\footnotesize
\setlength{\tabcolsep}{3pt}
\renewcommand{\arraystretch}{1}
\caption{\textbf{Quantitative comparison on DL3DV with 8 input views}. 
The pose-needed column indicates whether a method requires camera parameters. 
Metrics are PSNR$\uparrow$, SSIM$\uparrow$, and LPIPS$\downarrow$. 
Among feed-forward approaches, Ours requires neither intrinsics nor extrinsics and achieves the highest PSNR/SSIM and the lowest LPIPS.
The \textbf{best} and \underline{second-best} results are highlighted.}
\label{tab:dl3dv}
\resizebox{.85\linewidth}{!}{%
\begin{tabular}{Y C{0.75cm} *{3}{C{1cm}}}
\toprule
\multirow{2}{*}{\textbf{Method}} & \multirow{2}{*}{\textbf{Pose?}} &
\multicolumn{3}{c}{\textbf{Metrics @ 8 views}} \\
\cmidrule(lr){3-5}  &
& PSNR $\uparrow$ & SSIM $\uparrow$ & LPIPS $\downarrow$  \\
\midrule
PixelSplat \cite{charatan2024pixelsplat} & \cmark & 22.55 & 0.727 & 0.192   \\
MVSplat \cite{chen2024mvsplat}  & \cmark & 22.08 & 0.717 & 0.189  \\
FLARE \cite{zhang2025flare} & \xmark & 23.33 & 0.746 & 0.237 \\
AnySplat \cite{jiang2025anysplat}  & \xmark & 23.76 & 0.762 & 0.187 \\
\textbf{IFSplat(Ours)}  & \xmark & \underline{24.69} & \underline{0.809} & \underline{0.171} \\
\textbf{GIFSplat(Ours)} & \xmark & \textbf{24.91} & \textbf{0.824} & \textbf{0.164} \\
\bottomrule
\end{tabular}}
\end{table}

\paragraph{Generalization to out-of-domain datasets}

~\cref{tab:ood} results of cross-dataset evaluation of baselines, indicate that our iterative residual feed-forward design generalizes well to unconstrained captures, complementing the out-of-domain evaluation on DTU dataset, with over 2 dB performance gain. In addition, we further show the generalization ability of our model in Appendix, demonstrating that the robustness of our model even for in-the-wild images. 



\begin{table}[t]
\centering
\caption{\textbf{Out-of-distribution performance comparison}. Our method shows superior performance when cross-dataset evaluation on DTU using the model solely trained on RealEstate10K.
The \textbf{best} and \underline{second-best} results are highlighted.}
\label{tab:ood}
\resizebox{0.85\linewidth}{!}{%
\begin{tabular}{@{}lccccccc@{}}
\toprule
\multirow{2}{*}{\textbf{Methods}} & \multirow{2}{*}{\textbf{Pose?}} & \multicolumn{3}{c}{\textbf{DTU @ 2 views}} \\
\cmidrule(lr){3-5} 
& & \textbf{PSNR} $\uparrow$ & \textbf{SSIM} $\uparrow$ & \textbf{LPIPS} $\downarrow$  \\
\midrule
pixelSplat \cite{charatan2024pixelsplat} & \cmark & 11.551 & 0.321 & 0.633  \\
MVSplat \cite{chen2024mvsplat} & \cmark & 13.929 & 0.474 & 0.385  \\
NopoSplat \cite{ye2024no} & \xmark  & 17.899 & 0.629 & 0.279  \\
FLARE \cite{zhang2025flare} & \xmark   & 17.528 & 0.596 & 0.283  \\
AnySplat \cite{jiang2025anysplat} & \xmark  & 18.122 & 0.632 & 0.276  \\
\textbf{IFSplat(ours)}  & \xmark   & \underline{19.921} & \underline{0.701} & \underline{0.274}  \\
\textbf{GIFSplat(ours)}   & \xmark   & \textbf{20.214} & \textbf{0.716} & \textbf{0.251}  \\
\bottomrule
\end{tabular}%
}
\end{table}

\subsection{Ablation Study}
\cref{tab:ablation} validates the contribution of each component. 
Removing the iterative refinement (\emph{\wo Stage~2}) yields the largest degradation across all metrics, confirming the necessity of iterative residual updates.
Disabling the generative prior (\emph{\wo G.~prior}) also reduces performance, particularly in LPIPS, showing that the prior-aware cues effectively suppress artifacts and improve perceptual fidelity.
Disabling the window attention (\emph{\wo window}) also make substantial performance drop, which indicates the function of iteraction in 3D point space.
The full configuration achieves the best performance, demonstrating the complementarity between all components.
\newcolumntype{Y}{>{\raggedright\arraybackslash}p{2.1cm}}
\begin{table}[htbp]
\centering
\footnotesize
\setlength{\tabcolsep}{3pt}
\renewcommand{\arraystretch}{1}
\caption{\textbf{Ablation study. } We ablate the iterative refinement module (\wo Refinement), the window attention (\wo window att.), and the generative prior (\wo Gen. prior) from our full model, respectively. The full configuration achieves the best performance across all metrics.
The \textbf{best} and \underline{second-best} results are highlighted.}
\label{tab:ablation}
\resizebox{.85\columnwidth}{!}{%
\begin{tabular}{Y C{1.3cm} C{1.3cm} C{1.3cm}}
\toprule
\multirow{2}{*}{\textbf{Components}} &
\multicolumn{3}{c}{\textbf{Metrics}} \\
 \cmidrule(lr){2-4} 

& \textbf{PSNR} $\uparrow$ & \textbf{SSIM} $\uparrow$ & \textbf{LPIPS} $\downarrow$ \\
\midrule
\wo Refinement  & 24.781 & 0.826 & 0.169 \\
\wo window att.     & 25.327 & 0.837 & 0.152 \\
\wo Gen. Prior  & \underline{26.291} & \underline{0.854} & \underline{0.145} \\
\textbf{Full} & \textbf{26.559} & \textbf{0.867} & \textbf{0.138}\\
\bottomrule
\end{tabular}}
\end{table}

\subsection{Additional Experiments}
\paragraph{Hyperparameter analysis for iterative steps}
\label{App:hyper}

We vary the number of forward refinements $N$ while keeping all other settings fixed. All experiments are implemented on RealEstate10K dataset in a two-view setting. 
Accuracy improves monotonically from $N{=}0$ and exhibits clear diminishing returns after a few steps. 
In practice we use a small $N=3$ for all main experiments as a favorable trade-off between accuracy and latency .

\begin{table}[htbp]
\centering
\footnotesize
\setlength{\tabcolsep}{3pt}
\renewcommand{\arraystretch}{1}
\caption{\textbf{Analysis of iterative steps.}  We report metrics as the number of refinement steps $T$ increases from the initial prediction, for both IFSplat and our GIFSplat. Performance improves steadily and saturates around 3 iterations.}
\label{tab:ablation}
\resizebox{.95\columnwidth}{!}{%
\begin{tabular}{c C{1.1cm} C{1.1cm} C{1.1cm} C{1.1cm} C{1.1cm} C{1.1cm}}
\toprule
\multirow{2}{*}{\textbf{Steps}} &
\multicolumn{3}{c}{\textbf{Metrics@IFSplat}} & \multicolumn{3}{c}{\textbf{Metrics@GIFSplat}} \\
\cmidrule(lr){2-4}   \cmidrule(lr){5-7} 

& \textbf{PSNR} $\uparrow$ & \textbf{SSIM} $\uparrow$ & \textbf{LPIPS} $\downarrow$ & \textbf{PSNR} $\uparrow$ & \textbf{SSIM} $\uparrow$ & \textbf{LPIPS} $\downarrow$ \\
\midrule
initial  & 24.901 & 0.831 & 0.164 & 24.901 & 0.831 & 0.164  \\
1     & 25.512 & 0.843  &  0.155 & 25.774 & 0.845  &  0.151 \\
2     & 25.873 & 0.847  & 0.150 & 26.107 & 0.849  & 0.147 \\
3     & \underline{26.291} & \underline{0.854} & \textbf{0.145} & \underline{26.559} & \textbf{0.867} & \underline{0.138} \\
4     & \textbf{26.311} & \textbf{0.855} & \textbf{0.146} & \textbf{26.561} & \underline{0.865} & \textbf{0.137} \\

\bottomrule
\end{tabular}}
\end{table}

\begin{figure}[htbp]
  \centering
  \includegraphics[width=.95\linewidth]{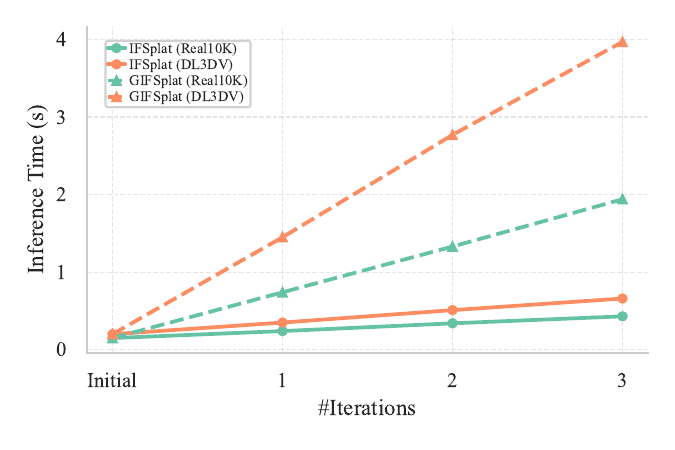}
  \caption{\textbf{Time complexity analysis.} Inference time of our methods IFSplat and GIFSplat on two datasets as a function of the number of refinement steps. Both methods scale approximately linearly with $T$.}
\label{fig:time}
\end{figure}

\paragraph{Additional experiments} 
Due to space constraints, we defer three sets of experiments to the Appendix: (1) a time-complexity analysis of our model, showing that our method retains second-scale runtime even with scene-specific refinement and fusion of generative priors, with Fig.~\ref{fig:time} illustrating one representative part of this analysis; (2) comparisons to 3D reconstruction baselines that incorporate generative priors, demonstrating that our method is consistently superior in both quality and efficiency in sparse-view settings; and (3) experiments on in-the-wild images demonstrating robustness under domain shift. Additional experimental settings, quantitative results, and qualitative visualizations are provided in the Appendix.

\section{Conclusion}
\label{sec:conclusion}


We presented GIFSplat, an iterative feed-forward 3D Gaussian Splatting framework that refines an initial prediction through multi-step forward-only residual updates guided by both observations and diffusion-enhanced cues. Our IFSplat baseline already brings scene-specific refinement without test-time backpropagation, and the full GIFSplat variant further injects a frozen generative prior via lightweight Gaussian-level discrepancy cues. Experiments on DL3DV, RealEstate10K, and DTU show consistent gains over recent feed-forward 3D reconstruction methods, especially in cross-domain settings.

\noindent\textbf{Limitations.}
Our refinement head still focus on static scenes and a small, fixed set of input modalities. Extending GIFSplat to dynamic content or to flexible geometric priors such as known depth maps or normal maps is an interesting direction for future work.

\section*{Acknowledgments}
This research was supported by the Australian Research Council Discovery Project (Grant Number DP240103334).

{
    \small
    \bibliographystyle{ieeenat_fullname}
    \bibliography{main}
}


\end{document}